\documentclass[conference]{IEEEtran}
\IEEEoverridecommandlockouts
\usepackage{cite}
\usepackage{amsmath,amssymb,amsfonts}
\usepackage{algorithmic}
\usepackage{graphicx}
\usepackage{textcomp}
\usepackage{xcolor}
\usepackage{booktabs}
\usepackage{multirow}
\usepackage{array}
\usepackage{ragged2e}
\usepackage{subcaption}
\usepackage[font=small,labelfont=bf]{caption}

\def\BibTeX{{\rm B\kern-.05em{\sc i\kern-.025em b}\kern-.08em
    T\kern-.1667em\lower.7ex\hbox{E}\kern-.125emX}}

\begin{document}

\title{Zero-Shot Captioning for Cultural Heritage: Automated Image
Analysis of Traditional Indonesian Clothing}

\author{
\IEEEauthorblockN{1\textsuperscript{st} Anugrah Aidin Yotolembah}
\IEEEauthorblockA{
Department of Informatics Engineering\\
Faculty of Computer Science\\
Brawijaya University\\
Malang, Indonesia\\
didiyotolembah@student.ub.ac.id
}
\and
\IEEEauthorblockN{2\textsuperscript{nd} Novanto Yudistira}
\IEEEauthorblockA{
Department of Informatics Engineering\\
Faculty of Computer Science\\
Brawijaya University\\
Malang, Indonesia\\
yudistira@ub.ac.id
}
\and
\IEEEauthorblockN{3\textsuperscript{rd} Gembong Edhi Setyawan}
\IEEEauthorblockA{
Department of Informatics Engineering\\
Faculty of Computer Science\\
Brawijaya University\\
Malang, Indonesia\\
gembong@ub.ac.id
}
}

\maketitle

\begin{abstract}
This paper presents Custom ZeroCLIP, a retrieval-augmented vision-language framework for zero-shot captioning of Indonesian traditional garments. The dataset contains 3,800 expert-annotated images from all 38 Indonesian provinces. Using a province-level inductive zero-shot protocol, the model is trained on 24 seen provinces, validated on 6 seen provinces, and evaluated on 8 unseen provinces. The framework combines a frozen CLIP ViT-B/32 image encoder, a CLIP text encoder, a BERT text encoder, and an LSTM caption decoder. During inference, unseen-province labels and captions are unavailable, and retrieval uses only captions from training provinces. No unseen-province image, label, or caption is used during training, validation, or retrieval-bank construction. Custom ZeroCLIP achieves a CLIPScore of 0.8536, BLEU-4 of 0.3342, and METEOR of 0.4859, outperforming existing baselines. Ablation results show that retrieval improves cultural vocabulary recovery with a 19.3\% METEOR gain, while human evaluation confirms stronger cultural accuracy and fluency. The results demonstrate the effectiveness of retrieval-augmented domain adaptation for culturally grounded caption generation in low-resource heritage settings. The dataset is publicly available at https://github.com/AnugrahAidinYotolembah/Traditional-Indonesian-Clothing-Captioning-Dataset.
\end{abstract}


\section{Introduction}
\label{sec:intro}

Indonesia's textile heritage, including Batik, Tenun Ikat, Ulos,
and Songket, reflects rich ethnolinguistic diversity and encodes
ceremonial as well as historical knowledge through motifs and
weaving practices \cite{b1,b3,b7,b16}. However, modernization has
gradually reduced these garments to ritual and museum contexts,
thereby endangering their cultural vocabulary \cite{b6}. Image
captioning provides a scalable approach for cultural preservation
through visually grounded textual descriptions \cite{b4,b8}.

Despite their strong performance, vision-language models such as
CLIP \cite{b14} and BLIP \cite{b18} remain limited in capturing
fine-grained Indonesian cultural terminology \cite{b24,b17}. This
motivates the problem of \emph{inductive zero-shot} cultural
captioning, where models are required to generate descriptions
for garments from unseen provinces without access to their
images, labels, or reference captions during training or
inference \cite{b26}.

In this work, inductive zero-shot is defined at the level of
province generalization. Unseen provinces are fully excluded from
training, validation, and retrieval-bank construction, while
inference-time retrieval is restricted to captions originating
only from seen provinces. Under this setting, we construct a
dataset consisting of 3,800 expert-annotated images spanning 38
Indonesian provinces, and propose \textbf{Custom ZeroCLIP}, a
retrieval-augmented vision-language framework that integrates
frozen CLIP encoders with a BERT–LSTM decoder for culturally
grounded caption generation. The proposed approach is evaluated
in terms of cultural accuracy, linguistic fluency, and zero-shot
generalization performance in a low-resource heritage setting.

We frame this work as cultural heritage documentation 
rather than generative art creation. Generated captions 
support museum digitization and cultural indexing, but 
require expert validation before deployment. Overall, 
results demonstrate effectiveness of retrieval-guided 
domain adaptation for culturally grounded image captioning.

\section{Related Work}

\subsection{Image Captioning and Vision-Language Pretraining}
Image captioning has evolved from CNN--RNN and attention-based
architectures \cite{b21,b22,b20} to transformer-based
vision-language models (VLMs) \cite{b33}. CLIP \cite{b14}
enables zero-shot transfer, while models such as BLIP
\cite{b18}, InstructBLIP, BLIP-2, Flamingo, PaLI, and Kosmos-2
improve multimodal reasoning through large-scale pretraining and instruction tuning \cite{b38}.

Recent VLMs such as BLIP-2, Flamingo, PaLI, and Kosmos-2 improve
multimodal reasoning through web-scale supervision, but their
Western-centric training data limits representation of
fine-grained cultural attributes in low-resource heritage domains
\cite{b39,b40}.

BERT and LSTM remain effective for low-resource captioning
\cite{b34,b35}. Our approach combines retrieval-guided decoding
and domain adaptation for cultural heritage captioning.

\subsection{Zero-Shot Learning for Visual Recognition}

Zero-shot learning (ZSL) generalizes to unseen classes without
labeled examples \cite{b26,b27}. Under the inductive setting
\cite{b26}, no unseen data or labels are available during
training, unlike the transductive variant. Retrieval-augmented methods extend ZSL by leveraging external memory at inference. Inspired by retrieval-augmented generation (RAG)
\cite{b19}, visual embeddings are matched with text 
embeddings, and the top-$K$ candidates are retrieved 
to guide caption generation without class labels 
\cite{b19}.

\subsection{Cultural Heritage AI and Research Gap}

Fashion captioning benchmarks such as DeepFashion and FashionIQ
are dominated by Western garments, limiting performance on
non-Western attire \cite{b12,b23}. Although prior work explores
cultural clothing in other regions, province-level datasets for
Indonesian attire remain scarce \cite{b8}. Existing approaches
focus mainly on recognition rather than cultural interpretation.
Our work addresses this gap by framing captioning as cultural
meaning reconstruction and introducing inductive zero-shot
captioning across all Indonesian provinces with
retrieval-augmented generation.

\section{Proposed Method}

\subsection{Problem Formulation}

Let $\mathcal{P} = \{p_1, \ldots, p_{38}\}$ denote the set of Indonesian provincial garment categories. The dataset is partitioned into 24 training provinces, 6 validation provinces, and 8 held-out unseen test provinces. The union of training and validation provinces forms the set of seen provinces $\mathcal{P}^{s}$, while the held-out test provinces form the unseen set $\mathcal{P}^{u}$.

Following the inductive zero-shot learning protocol \cite{b26,b27}, no image, label, or reference caption from $\mathcal{P}^{u}$ is used during training, validation, hyperparameter tuning, or retrieval-bank construction. The retrieval bank used during inference contains only captions from the training split.

\subsection{System Architecture}

\begin{figure*}[t]
  \centering
  \includegraphics[width=\textwidth]{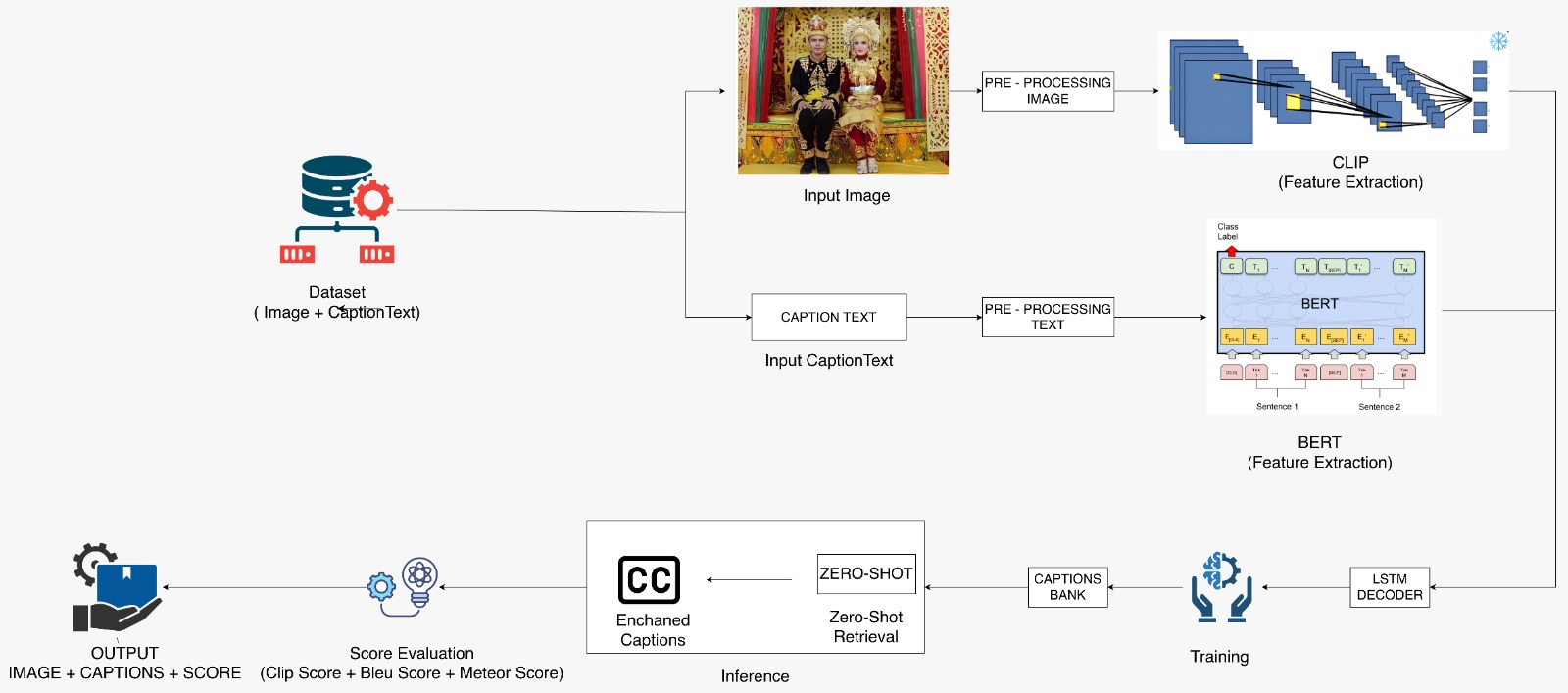}
  \caption{Overview of Custom ZeroCLIP. Training (left) optimizes a BERT-LSTM decoder using frozen CLIP embeddings from seen provinces. Inference (right) applies cosine-similarity retrieval to generate captions for unseen provinces without labels.}
  \label{fig:architecture}
\end{figure*}

As illustrated in Fig.~\ref{fig:architecture}, the system
operates in two phases. During the \textbf{training} phase,
paired image-caption data from $\mathcal{P}^{s}$ is used to
optimize a BERT-LSTM decoder conditioned on frozen CLIP visual
embeddings, with no CLIP parameters being modified. During the
\textbf{inference} phase on $\mathcal{P}^{u}$, a zero-shot
retrieval step first identifies culturally proximate caption
candidates from the training corpus. These candidates are then
used to condition the LSTM decoder for label-free caption
generation.

\begin{table}[htbp]
\caption{Model Components and Training Configuration}
\label{tab:components}
\centering
\small
\setlength{\tabcolsep}{4pt}
\begin{tabular}{lccc}
\hline
\textbf{Component} &
\textbf{Frozen} &
\textbf{Trainable} &
\textbf{Used at Inference} \\
\hline
CLIP Image Encoder & Yes & No & Yes \\
CLIP Text Encoder & Yes & No & Yes \\
BERT Encoder & No & Yes & Yes \\
Projection Layers & No & Yes & Yes \\
LSTM Decoder & No & Yes & Yes \\
\hline
\end{tabular}
\end{table}

\subsection{CLIP-Based Visual Encoding}

A frozen CLIP ViT-B/32 image encoder maps an input image $I$ into a visual embedding $\mathbf{v} = f_{\mathrm{CLIP-image}}(I)$. All captions $c_i$ in the retrieval bank are pre-encoded using a frozen CLIP text encoder to obtain embeddings $\mathbf{r}_i = f_{\mathrm{CLIP-text}}(c_i)$, for $i=1,\dots,N$. Both modalities share CLIP's aligned space, enabling direct retrieval via cosine similarity. The CLIP encoders remain frozen during training and inference to preserve pretrained alignment and prevent catastrophic forgetting.

\begin{figure}[t]
  \centering
  \includegraphics[width=\columnwidth]{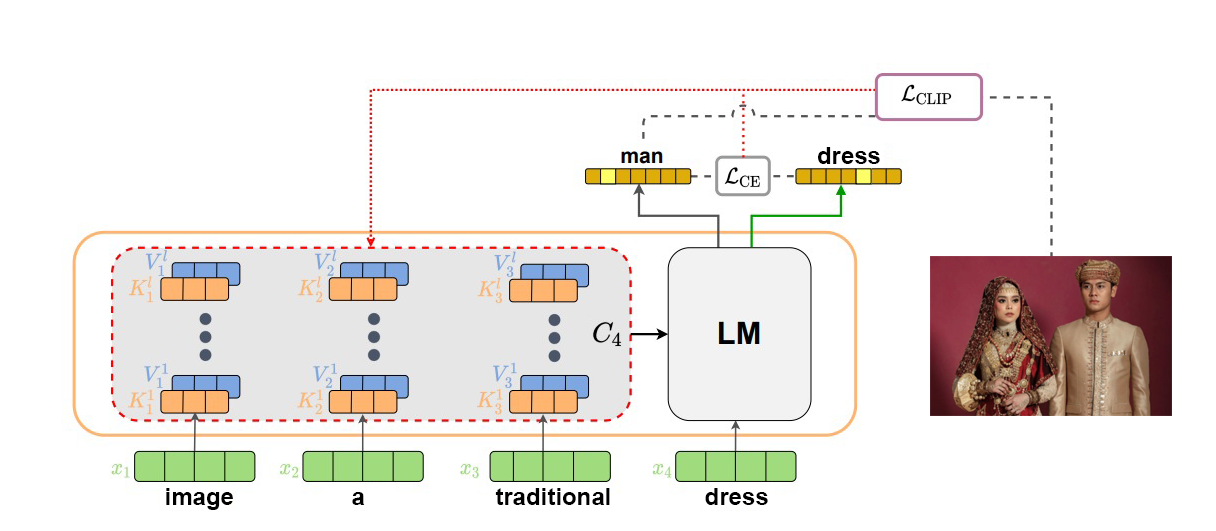}
  \caption{Training pipeline of Custom ZeroCLIP. Image and caption tokens are encoded by frozen CLIP, projected into the LM, and optimized using $\mathcal{L}_{CE}$ and $\mathcal{L}_{CLIP}$. The BERT encoder, projection layers, and LSTM decoder are trained while CLIP remains frozen.}
  \label{fig:pipeline}
\end{figure}

\subsection{BERT-LSTM Caption Decoder}

The decoder combines retrieved textual context and frozen CLIP visual features using a BERT encoder and an LSTM autoregressive decoder. For $K$ retrieved captions $c_i$ from the seen-province bank, each is encoded by a trainable BERT:
\begin{equation}
\mathbf{e}_i = f_{\mathrm{BERT}}(c_i), \quad i=1,\dots,K
\end{equation}

The embeddings are mean-pooled into a global context:
\begin{equation}
\bar{\mathbf{e}} = \frac{1}{K}\sum_{i=1}^{K}\mathbf{e}_i
\end{equation}

This text context is fused with the frozen CLIP image embedding $\mathbf{v}$ via learnable projections:
\begin{equation}
\mathbf{z} = W_v\mathbf{v} + W_t\bar{\mathbf{e}}
\end{equation}

The fused feature conditions an LSTM-based autoregressive decoder:
\begin{equation}
\mathbf{h}_t = \mathrm{LSTM}(\mathbf{h}_{t-1}, y_{t-1}, \mathbf{z}), \quad
\hat{y}_t = \mathrm{Softmax}(W_o \mathbf{h}_t + b)
\end{equation}

At inference, retrieved captions provide contextual guidance. The CLIP image encoder remains frozen, while training updates only the BERT, projection layers, and LSTM. For multi-subject images (e.g., couples), the frozen CLIP captures the global visual context. Retrieved captions supply explicit relational cues, guiding the LSTM to correctly attribute specific garments to respective subjects without requiring bounding boxes.

\subsection{Zero-Shot Retrieval at Inference}

During inference, the model has no access to unseen-
province labels or unseen reference captions. The 
retrieval bank is built solely from captions in the 
seen-province training split.

\textbf{Retrieval Parameter Selection:} Across all 
experiments, we empirically set $K=5$ based on 
validation performance on the seen-province validation 
split before evaluation on unseen test provinces. This 
setting balances retrieval specificity, computational 
efficiency, and noise reduction from excessive 
candidates while preserving relevant cultural context.

Given a test image embedding $\mathbf{v}$, cosine similarity is computed against all text embeddings from the retrieval bank as $s_i = (\mathbf{v}^{\top}\mathbf{r}_i) / (\|\mathbf{v}\|\|\mathbf{r}_i\|)$. The top-$K$ retrieved captions are selected via $C^* = \mathrm{TopK}(s_i)$.

The retrieved captions provide lexical and cultural cues for the
LSTM decoder while preserving the inductive zero-shot protocol,
since no unseen-province annotations are accessed.

\section{Experiments}

\subsection{Dataset and Implementation}

The \textit{Indonesian Traditional Attire Dataset} contains 3,800 images from 38 Indonesian provinces (100 images per province), collected from museum archives, cultural organizations, educational repositories, and professional photography sources. All samples are annotated by domain experts, including anthropologists and textile scholars \cite{b8}, with labels covering garment categories (e.g., \textit{kebaya}, \textit{songket}, and \textit{ulos}), ceremonial attributes, ethnic affiliation, and cultural context.

The dataset includes male, female, couple, ceremonial, and isolated garment images for fine-grained textile understanding. We adopt a province-level inductive zero-shot setting, where unseen provinces are fully excluded from training, validation, and retrieval-bank construction.

The province-level dataset split is summarized in Table~\ref{tab:province_split}, ensuring strict separation between training, validation, and unseen test provinces.

\begin{table}[t]
\caption{Province-Level Dataset Split}
\label{tab:province_split}
\centering
\footnotesize
\setlength{\tabcolsep}{3pt}

\begin{tabular}{p{1.8cm}p{5.1cm}}
\toprule
\centering\arraybackslash\textbf{Split} &
\centering\arraybackslash\textbf{Provinces} \\
\midrule

Training
& Bali, East Java, Central Java, West Java, Papua, West Papua,
East Kalimantan, South Kalimantan, Central Kalimantan,
Central Sulawesi, South Sulawesi, West Sulawesi, Gorontalo,
Lampung, Bengkulu, Riau Islands, Riau, Jakarta, Yogyakarta,
West Nusa Tenggara, Maluku, North Kalimantan,
West Kalimantan, Central Papua \\
\midrule

Validation
& Southwest Papua, South Papua, Highland Papua, North Sulawesi,
South Central Java, North East Kalimantan \\
\midrule

Unseen Test
& Aceh, Banten, Jambi, North Maluku, East Nusa Tenggara,
Southeast Sulawesi, South Sumatra, North Sumatra \\

\bottomrule
\end{tabular}
\end{table}

\begin{figure}[t]
\centering

%================ ROW 1 ================
\begin{subfigure}{0.31\columnwidth}
    \centering
    \includegraphics[width=\linewidth,height=2.6cm,keepaspectratio]{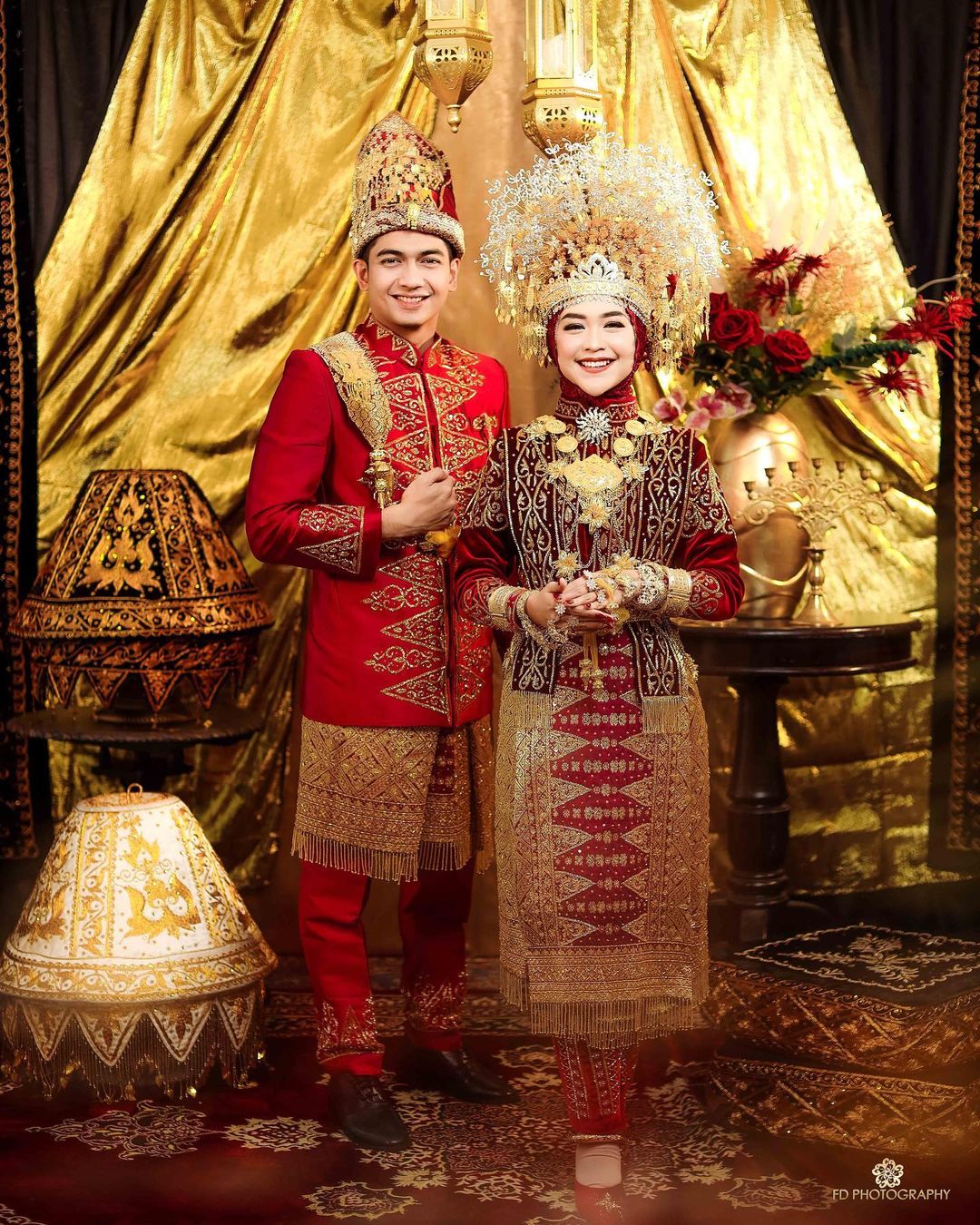}
    
    {\scriptsize \textbf{Aceh} (Unseen)}
\end{subfigure}
\hfill
\begin{subfigure}{0.31\columnwidth}
    \centering
    \includegraphics[width=\linewidth,height=2.6cm,keepaspectratio]{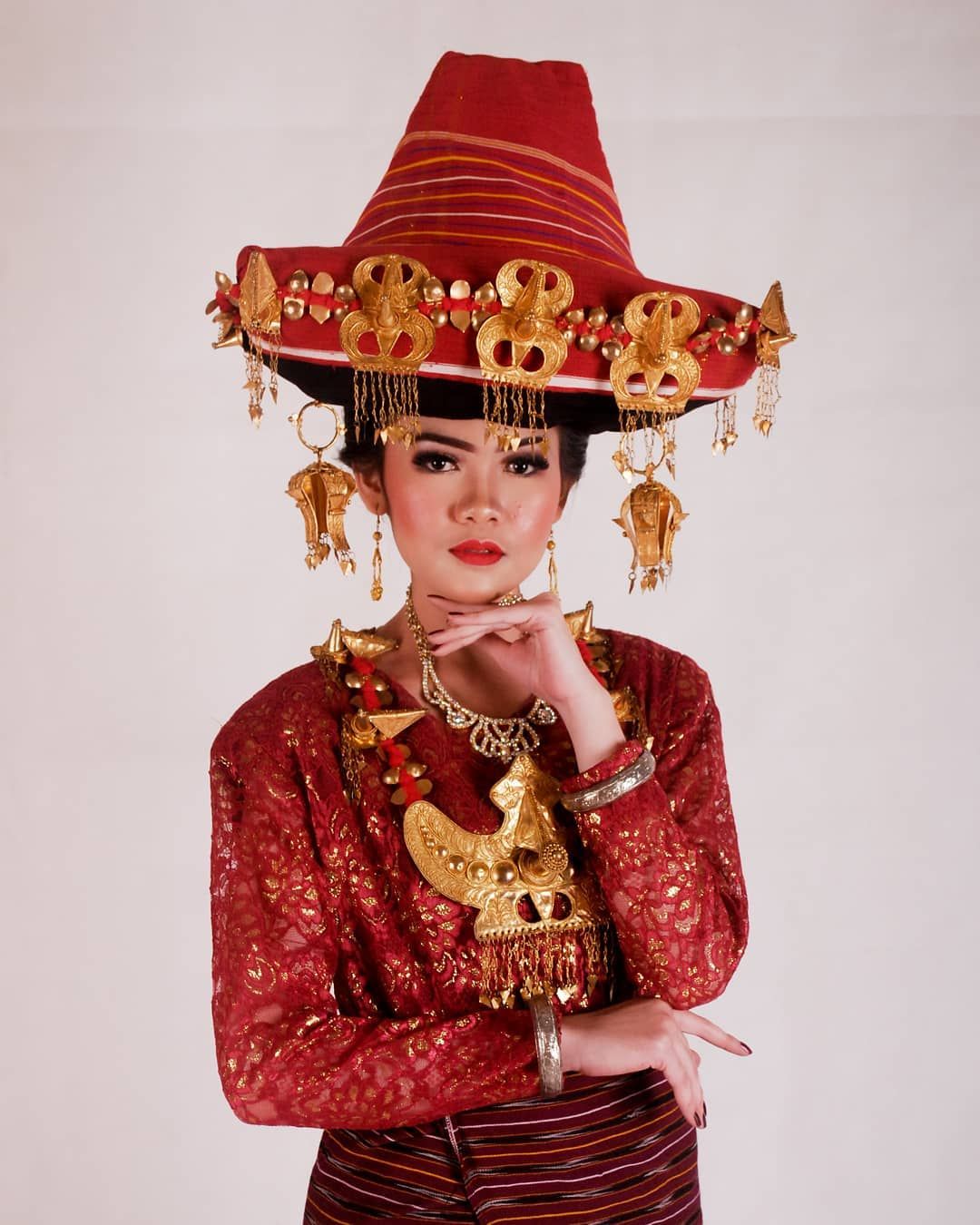}
    
    {\scriptsize \textbf{N. Sumatra} (Unseen)}
\end{subfigure}
\hfill
\begin{subfigure}{0.31\columnwidth}
    \centering
    \includegraphics[width=\linewidth,height=2.6cm,keepaspectratio]{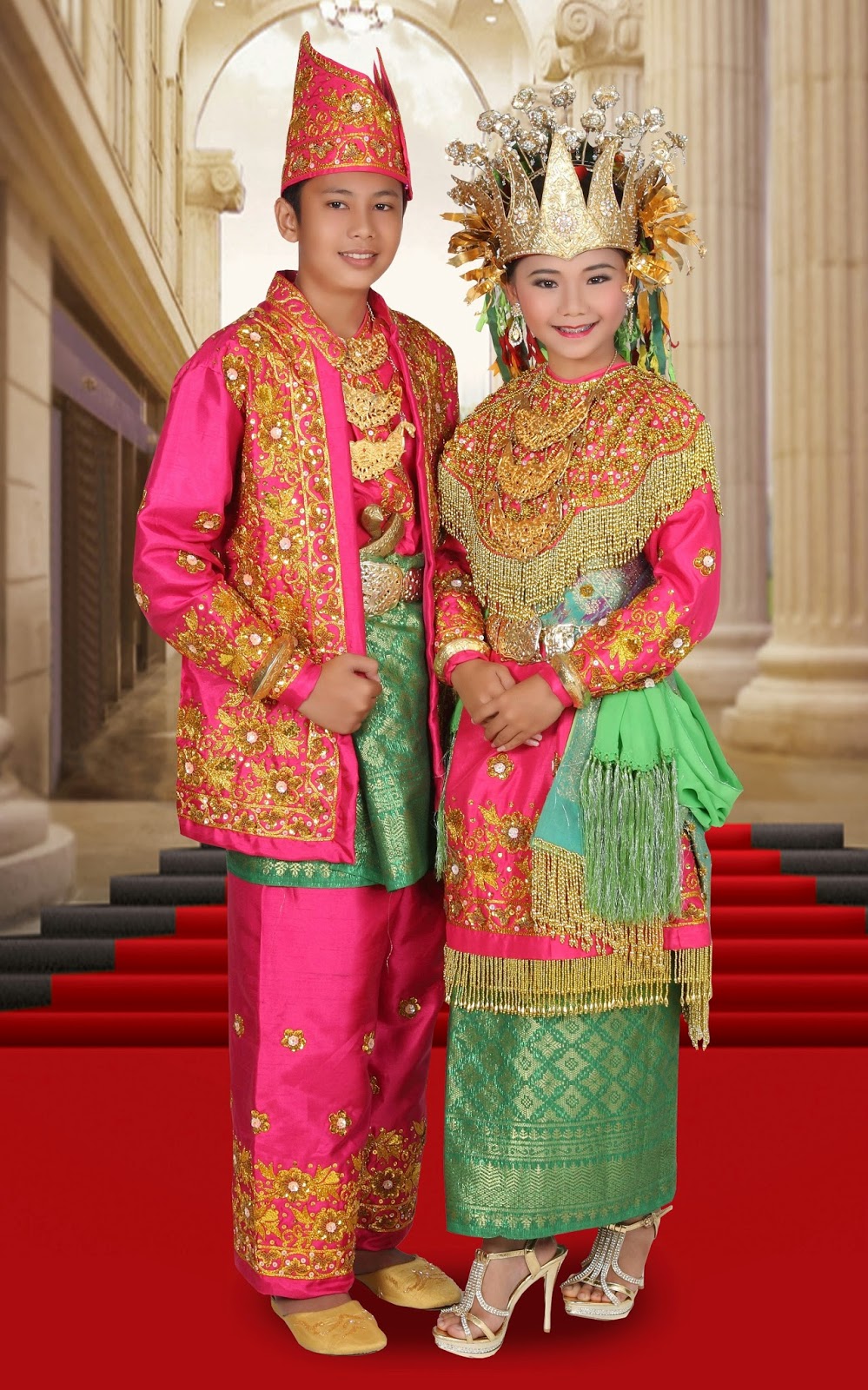}
    
    {\scriptsize \textbf{Jambi} (Unseen)}
\end{subfigure}

\vspace{0.4em}

%================ ROW 2 ================
\begin{subfigure}{0.31\columnwidth}
    \centering
    \includegraphics[width=\linewidth,height=2.6cm,keepaspectratio]{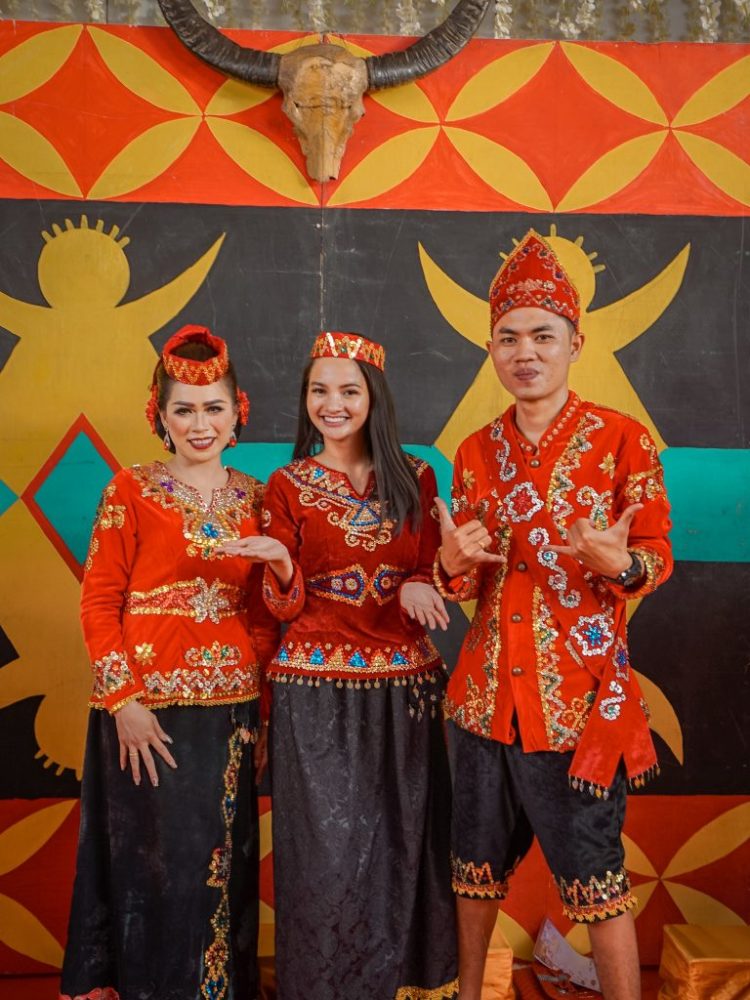}
    
    {\scriptsize \textbf{C. Sulawesi} (Seen)}
\end{subfigure}
\hfill
\begin{subfigure}{0.31\columnwidth}
    \centering
    \includegraphics[width=\linewidth,height=2.6cm,keepaspectratio]{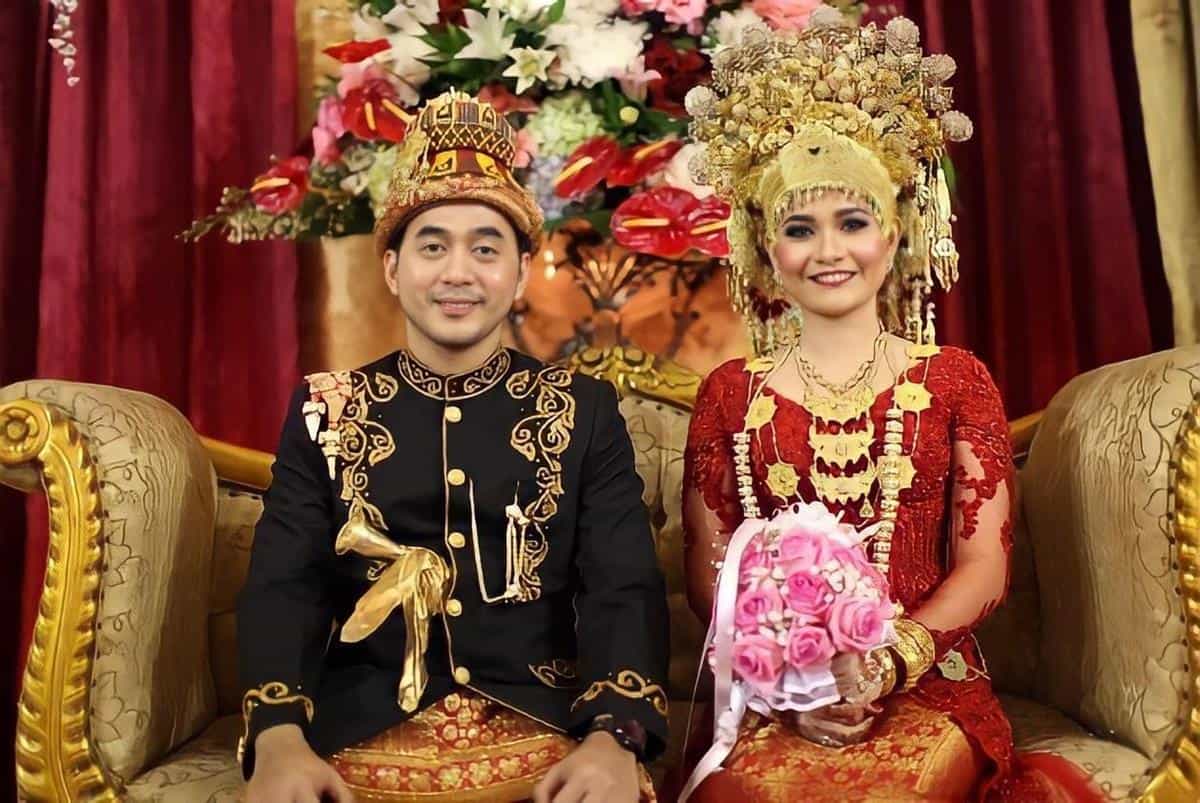}
    
    {\scriptsize \textbf{E. Java} (Seen)}
\end{subfigure}
\hfill
\begin{subfigure}{0.31\columnwidth}
    \centering
    \includegraphics[width=\linewidth,height=2.6cm,keepaspectratio]{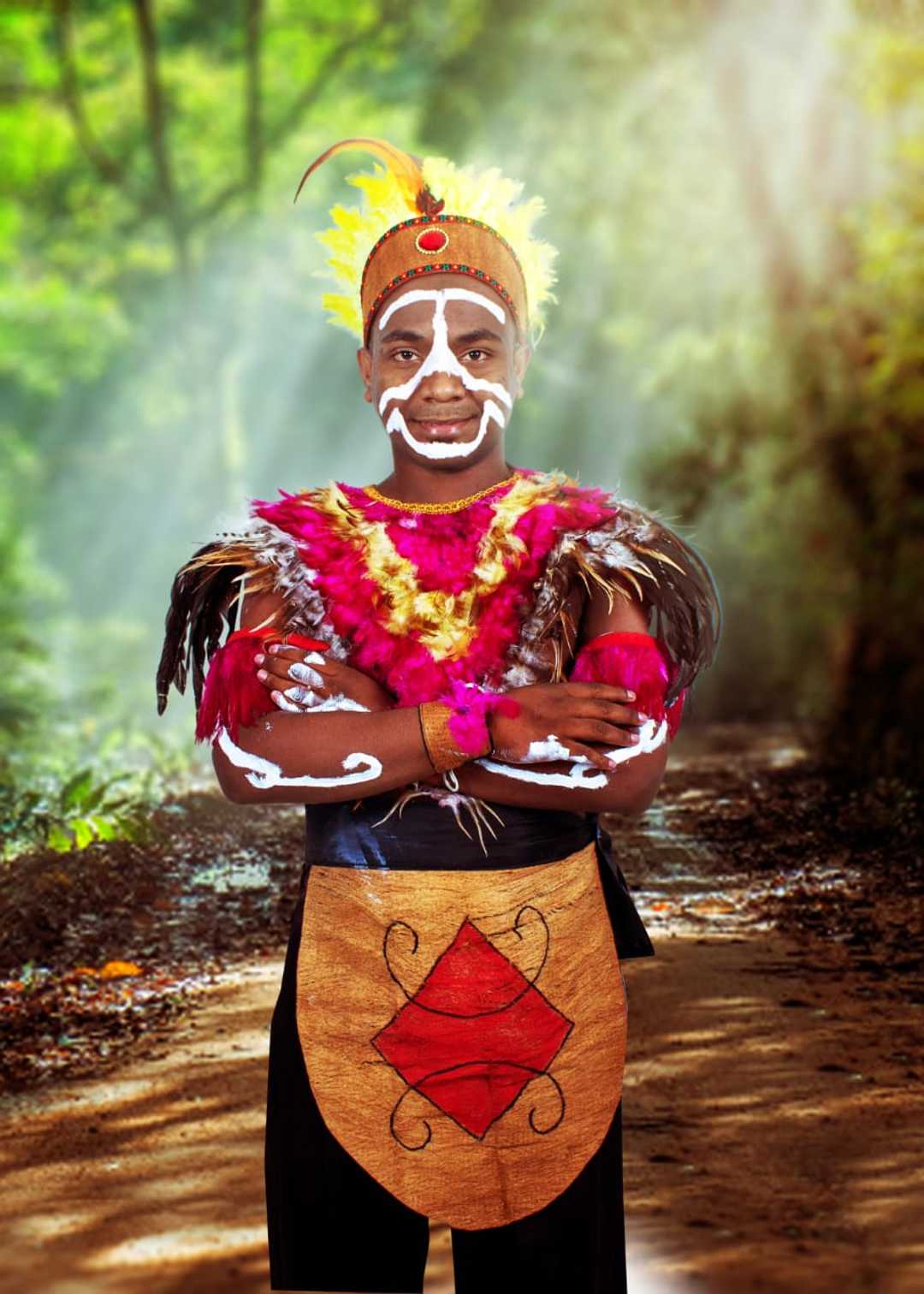}
    
    {\scriptsize \textbf{Papua} (Seen)}
\end{subfigure}

\vspace{0.3em}

\caption{Representative samples from seen and unseen provinces used in the inductive zero-shot evaluation protocol \cite{b37}.}
\label{fig:samples}

\end{figure}

Fig.~\ref{fig:samples} illustrates the visual diversity of Indonesian attire under the zero-shot protocol \cite{b37}, with full 38-province visualizations provided in the supplementary material due to page constraints.

All images are resized to $224 \times 224$ and normalized using ImageNet statistics \cite{b6,b19}. Data augmentation includes random horizontal flipping ($p=0.5$), color jittering, and rotations up to $15^\circ$ \cite{b22}. Captions are tokenized with the BERT tokenizer (maximum length: 512 tokens), while synonym replacement and back-translation improve robustness to cultural terminology \cite{b4,b15,b35}.

\begin{figure}[t]
\centering
\includegraphics[width=\columnwidth]{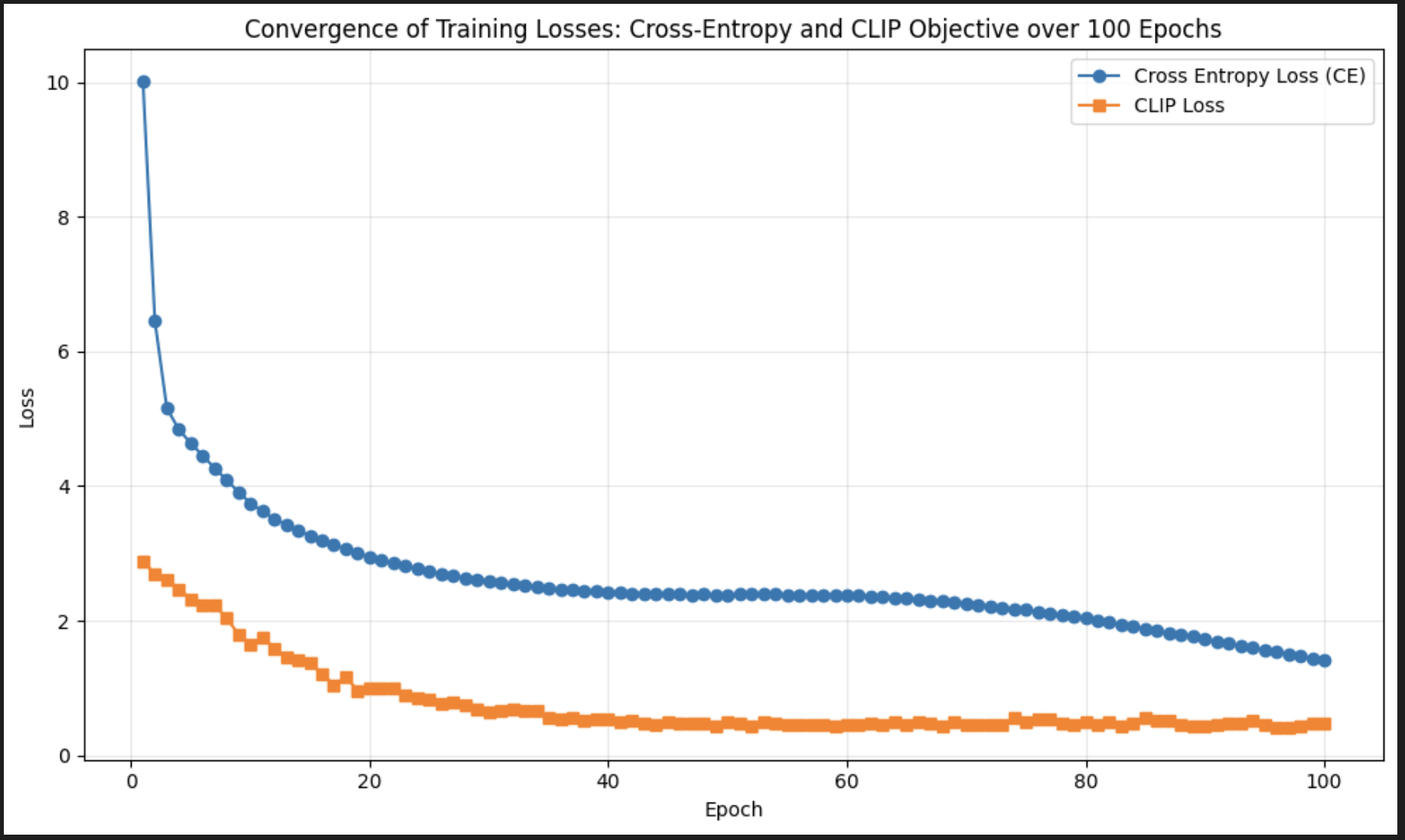}
\caption{Training and validation loss curves showing stable convergence without overfitting.}
\label{fig:loss}
\end{figure}

The CLIP ViT-B/32 encoder remains frozen during training, while the BERT encoder, projection layers, and LSTM decoder are jointly optimized using AdamW ($2 \times 10^{-5}$, 100 epochs). Fig.~\ref{fig:loss} shows stable convergence without overfitting.

\subsection{Quantitative Results}

Table~\ref{tab:results} reports results on the inductive
zero-shot test split comprising 8 unseen provinces. Under the
current setting, Custom ZeroCLIP achieves the best performance
across all metrics.

Compared with the strongest baseline per metric, Custom
ZeroCLIP improves CLIPScore by $+$1.97\% over InstructBLIP
(0.8371), BLEU-4 by $+$18.64\% over the CLIP baseline (0.2817),
and METEOR by $+$10.18\% over InstructBLIP (0.4410). The
BLEU-4 and METEOR gains indicate improved recovery of
province-specific terminology such as \textit{kebaya},
\textit{songket}, \textit{Meukutop}, and
\textit{blangkon} through retrieval-guided domain adaptation.

We acknowledge an asymmetric comparison since the proposed framework is domain-adapted while baseline VLMs are evaluated off-the-shelf. Fine-tuning foundational models presents significant computational barriers in this low-resource setting. Thus, baselines are included primarily to highlight their limitations on fine-grained cultural terminologies. Reported gains reflect the effectiveness of retrieval-augmented domain adaptation via a lightweight LSTM, rather than architectural superiority.

\begin{table}[htbp]
\caption{
Zero-shot performance on 8 unseen provinces.
Bold indicates the best result in each column.
$^\dagger$ Off-the-shelf evaluation without domain fine-tuning.
$^\ddagger$ Retrieval disabled during inference.
}
\label{tab:results}

\centering
\small
\setlength{\tabcolsep}{5pt}

\begin{tabular}{lccc}
\hline
\centering\arraybackslash\textbf{Model} &
\centering\arraybackslash\textbf{CLIPScore} &
\centering\arraybackslash\textbf{BLEU-4} &
\centering\arraybackslash\textbf{METEOR} \\
\hline

CLIP Baseline$^\dagger$
& 0.8021 & 0.2817 & 0.1274 \\

BLIP$^\dagger$
& 0.7729 & 0.2100 & 0.2881 \\

InstructBLIP$^\dagger$
& 0.8371 & 0.2353 & 0.4410 \\

MSGIT Base$^\dagger$
& 0.6932 & 0.2530 & 0.2730 \\

\hline

ZeroCLIP w/o Retrieval$^\ddagger$
& 0.8214 & 0.2743 & 0.3921 \\

Retrieval-only Top-1$^\ddagger$
& 0.8104 & 0.3187 & 0.4201 \\

\textbf{Custom ZeroCLIP}
& \textbf{0.8536}
& \textbf{0.3342}
& \textbf{0.4859} \\

\hline
\end{tabular}

\vspace{0.3em}

\footnotesize
$^\dagger$ Off-the-shelf models evaluated without domain-specific fine-tuning. \\
$^\ddagger$ Retrieval-only Top-1 directly outputs the nearest-neighbor caption without LSTM generation.

\end{table}

\begin{table*}[!t]
\centering
\small
\setlength{\tabcolsep}{4pt}
\renewcommand{\arraystretch}{1.04}

\resizebox{\textwidth}{!}{%
\begin{tabular}{
  >{\centering\arraybackslash}m{2.5cm}
  >{\centering\arraybackslash}m{2.7cm}
  >{\RaggedRight\arraybackslash}m{7.5cm}
  >{\RaggedRight\arraybackslash}m{7.5cm}
}
\toprule

\textbf{Image} &
\textbf{Model} &
\multicolumn{1}{c}{\textbf{Generated Caption}} &
\multicolumn{1}{c}{\textbf{Reference Caption}} \\

\midrule

\multirow{6}{*}{
\begin{tabular}{c}
\includegraphics[height=2.0cm,keepaspectratio]{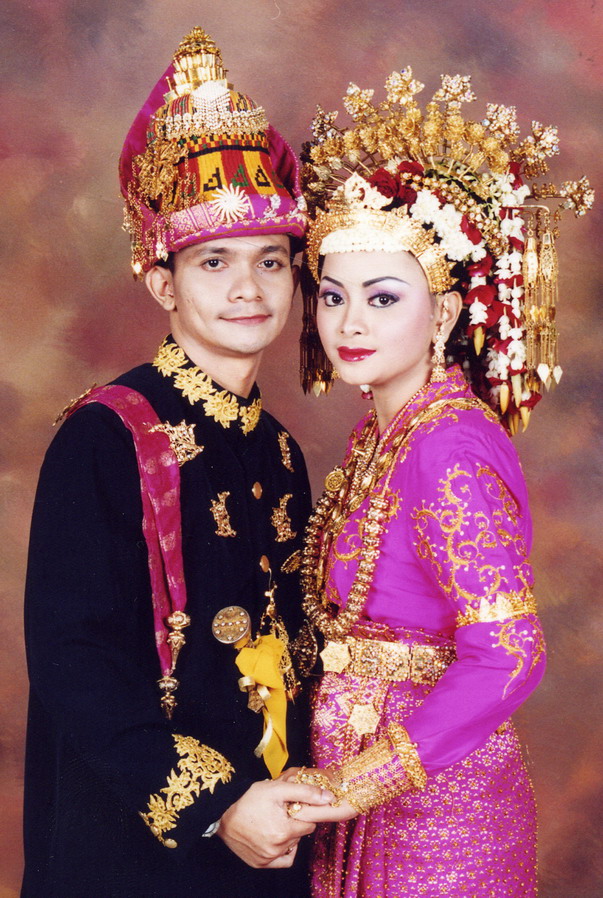} \\
{\scriptsize \textbf{Aceh} (Unseen)}
\end{tabular}}

& CLIP Baseline
& A person wearing traditional red clothing with gold accessories.
& \multirow{6}{7.5cm}{\RaggedRight
A couple wearing traditional Acehnese wedding attire. The groom
wears Linto Baro with gold embroidery and a Meukutop headdress.
The bride wears a violet kebaya, gold songket, and Suntiang
Aceh, reflecting Acehnese ceremonial heritage.
\textit{(Province unseen during training.)}}
\\[0.2em]

& BLIP
& A couple in matching red outfits posing for a photo.
& \\[0.2em]

& ZeroCLIP
& Image of a couple wearing traditional Indonesian wedding
clothes.
& \\[0.2em]

& InstructBLIP
& Two people wearing elaborate red and gold ceremonial costumes.
& \\[0.2em]

& MSGIT Base
& A man and a woman in traditional dress.
& \\[0.2em]

& \textbf{Custom ZeroCLIP}
& Image of a couple in Acehnese wedding attire, groom in Linto
Baro with Meukutop, bride in violet kebaya, songket, and
Suntiang Aceh headdress.
& \\

\midrule

\multirow{6}{*}{
\begin{tabular}{c}
\includegraphics[height=2.0cm,keepaspectratio]{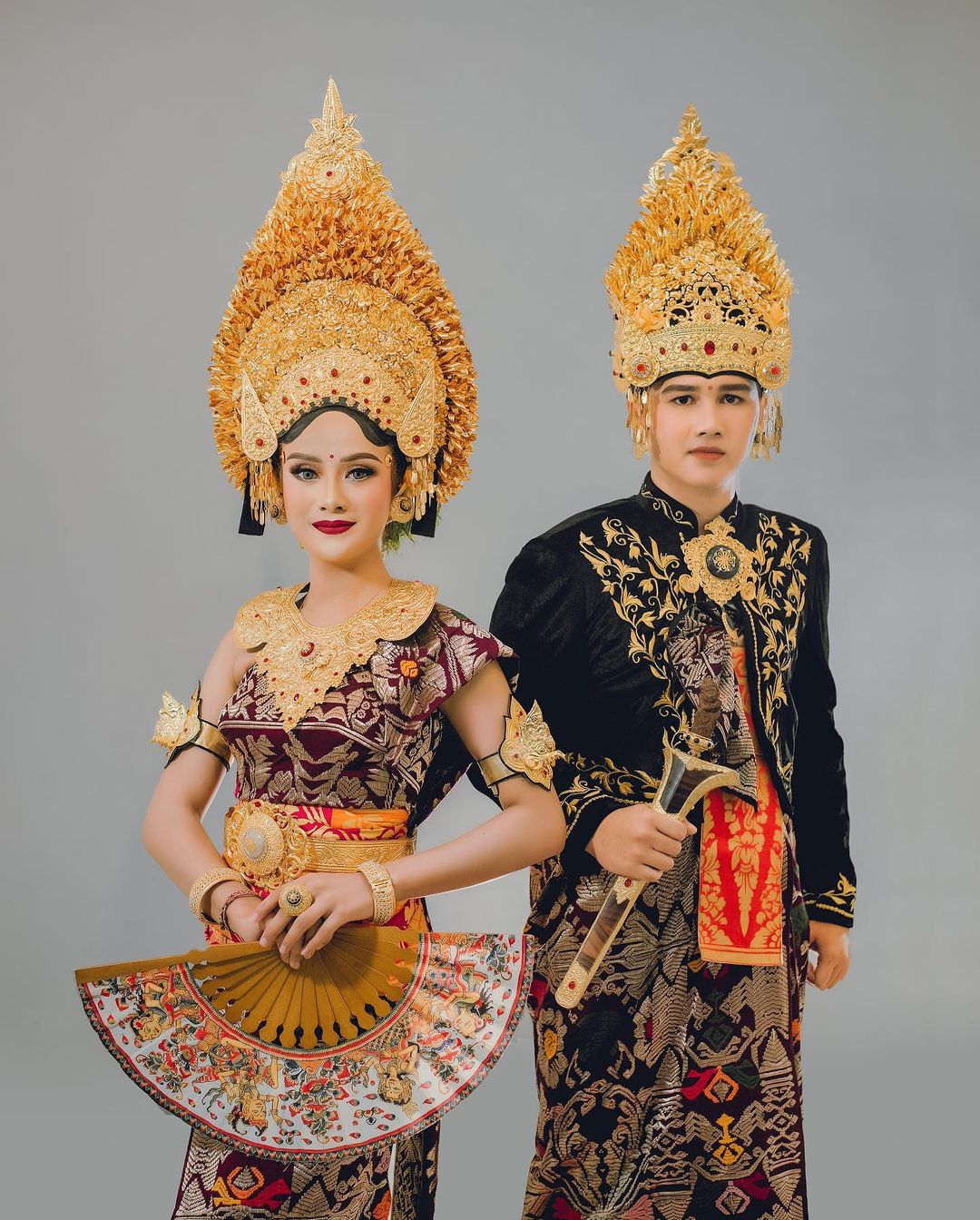} \\
{\scriptsize \textbf{Bali} (Seen)}
\end{tabular}}

& CLIP Baseline
& A woman in a colorful traditional dress with a headdress.
& \multirow{6}{7.5cm}{\RaggedRight
Balinese temple attire with kebaya lace, kamen batik cloth,
selendang sash, and golden floral headpiece representing
Hindu-Balinese spiritual identity.
\textit{(Province seen during training.)}}
\\[0.2em]

& BLIP
& A woman wearing a white lace top and batik skirt.
& \\[0.2em]

& ZeroCLIP
& Image of a woman in traditional clothing.
& \\[0.2em]

& InstructBLIP
& A person wearing traditional Southeast Asian clothing with
flowers.
& \\[0.2em]

& MSGIT Base
& A woman in traditional dress.
& \\[0.2em]

& \textbf{Custom ZeroCLIP}
& Image of Balinese temple attire with kebaya lace, kamen batik,
selendang sash, and golden floral headpiece.
& \\

\midrule

\multirow{6}{*}{
\begin{tabular}{c}
\includegraphics[height=2.0cm,keepaspectratio]{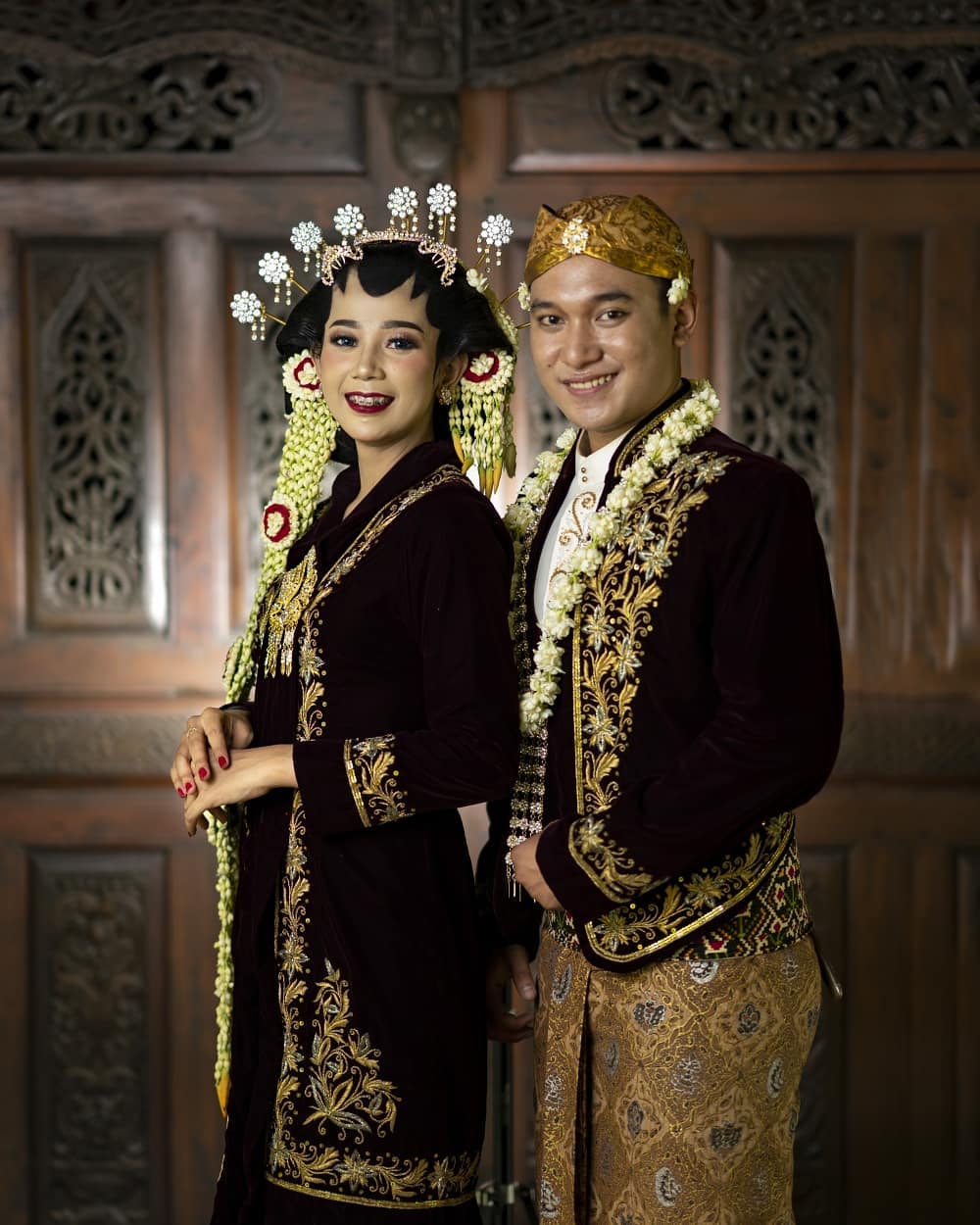} \\
{\scriptsize \textbf{East Java} (Seen)}
\end{tabular}}

& CLIP Baseline
& A person wearing a batik shirt and traditional hat.
& \multirow{6}{7.5cm}{\RaggedRight
East Javanese wedding attire featuring beskap jacket, blangkon
headpiece, jarik batik cloth, and keris dagger, symbolizing
Javanese nobility and ceremonial heritage.
\textit{(Province seen during training.)}}
\\[0.2em]

& BLIP
& A man and woman in traditional Javanese wedding costume.
& \\[0.2em]

& ZeroCLIP
& Image of a couple in traditional Javanese wedding attire.
& \\[0.2em]

& InstructBLIP
& Two people in elaborate traditional Indonesian wedding outfits.
& \\[0.2em]

& MSGIT Base
& Man and woman in traditional dress.
& \\[0.2em]

& \textbf{Custom ZeroCLIP}
& Image of East Javanese wedding attire with beskap jacket,
blangkon headpiece, jarik batik cloth, and keris dagger.
& \\

\bottomrule
\end{tabular}}

\caption{Qualitative comparison across representative garment
categories from seen and unseen provinces. Baseline VLMs
generate visually plausible but culturally generic captions,
whereas Custom ZeroCLIP recovers province-specific terminology
through retrieval-guided zero-shot decoding.}
\label{tab:qualitative}
\end{table*}

\subsection{Data Leakage Control}

To preserve inductive zero-shot validity, retrieval is restricted to training-split captions only, with no image, label, or reference caption from the 8 unseen provinces included. Near-duplicate filtering using CLIP image similarity (threshold 0.95) further prevents test overlap.

To distinguish province-level from vocabulary-level zero-shot, cultural term coverage is analyzed across train/val/test splits. Terms such as \textit{kebaya}, \textit{songket}, \textit{Meukutop}, \textit{Linto Baro}, and \textit{Suntiang Aceh} are tracked to determine whether recovery reflects \textbf{retrieval-guided transfer} from seen captions or genuine unseen-token generalization.

\subsection{Ablation Study}
\label{sec:ablation}

Table~\ref{tab:ablation} isolates the contributions of retrieval and decoding. Disabling retrieval causes a significant METEOR drop ($-19.3\%$). While domain adaptation maintains a moderate CLIPScore, retrieval proves essential for recovering fine-grained cultural vocabulary.

To verify the LSTM's generative value, we compare against a Retrieval-only Top-1 baseline that outputs the nearest caption without decoding. Custom ZeroCLIP outperforms this baseline across all metrics. Furthermore, modest n-gram overlap (23\% unigram, 8\% bigram) with the top-1 caption confirms genuine culturally-grounded generation rather than pure copying.

\begin{table}[htbp]
\caption{Ablation Study on Retrieval Mechanism}
\label{tab:ablation}
\centering
\small
\setlength{\tabcolsep}{5pt}
\begin{tabular}{lccc}
\toprule
\textbf{Model Variant} & \textbf{CLIPScore} & \textbf{BLEU-4} & \textbf{METEOR} \\
\midrule
ZeroCLIP w/o Retrieval & 0.8214 & 0.2743 & 0.3921 \\
Retrieval-only Top-1 & 0.8104 & 0.3187 & 0.4201 \\
\textbf{Custom ZeroCLIP} & \textbf{0.8536} & \textbf{0.3342} & \textbf{0.4859} \\
\bottomrule
\end{tabular}
\end{table}

\subsection{Qualitative Analysis}

Table~\ref{tab:qualitative} compares captions across three
representative garment categories, including the unseen Acehnese
case. General-purpose baselines generate visually plausible but
culturally generic descriptions without provincial identifiers
or ceremonial terminology.

Despite receiving no training data from Aceh, Custom ZeroCLIP
recovers province-specific terms such as \textit{Linto Baro},
\textit{Meukutop}, \textit{kebaya}, \textit{songket}, and
\textit{Suntiang Aceh}, which are absent from baseline outputs.
This improvement is enabled by retrieval-guided decoding using
captions from culturally related seen provinces. The Balinese
and East Javanese examples further demonstrate consistent
generation of garment names and symbolic accessories such as
kamen batik, jarik batik, selendang, keris dagger, and
blangkon headpiece.

\subsection{Human Evaluation}

Five knowledgeable annotators (two textile researchers, one museum archivist, two graduate researchers) assessed caption quality using a stratified random sample of 80 images from 8 unseen provinces (10 per province). Each evaluated all 80 images in a blind setting. Evaluation used a 1--5 Likert scale for \textit{cultural accuracy} (province-specific terminology, ceremonial context) and \textit{fluency} (grammar, coherence), reporting mean $\pm$ std dev and Fleiss' $\kappa$ for inter-annotator agreement.

For practicality, human evaluation focuses on the strongest
baseline models and the proposed framework, following common
practice in vision-language evaluation.

\begin{table}[htbp]
\caption{Human Evaluation Results on Unseen Provinces}
\label{tab:human_eval}
\centering
\small
\setlength{\tabcolsep}{6pt}
\begin{tabular}{lccc}
\toprule
\textbf{Model} & \textbf{Accuracy} & \textbf{Fluency} & \textbf{Kappa} \\
\midrule
BLIP                    & $3.1 \pm 0.8$ & $3.8 \pm 0.6$ & 0.42 \\
InstructBLIP            & $3.6 \pm 0.9$ & $4.1 \pm 0.5$ & 0.51 \\
\textbf{Custom ZeroCLIP} & \textbf{$4.4 \pm 0.5$} & \textbf{$4.3 \pm 0.4$} & \textbf{0.68} \\
\bottomrule
\end{tabular}
\end{table}

Custom ZeroCLIP achieves the highest scores in both evaluation
criteria, indicating improved cultural grounding and fluent
caption generation compared to the strongest baseline models.
Scores above 4.0 indicate strong perceived caption quality.

These findings are consistent with the quantitative evaluation,
suggesting that retrieval-augmented domain adaptation improves
perceived cultural relevance under the current experimental
setting, particularly for unseen provincial garments.

\subsection{Discussion and Limitations}
Results show zero-shot retrieval enables culturally coherent captioning. From an AIART perspective, this framework moves beyond object recognition to \textit{cultural meaning reconstruction}. By generating ceremonial terminologies, it aids the artistic interpretation of traditional textiles, bridging visual motifs with historical narratives. However, since retrieval systems may oversimplify cultural nuances, expert validation remains essential. Generated captions should therefore be treated as \textbf{assistive rather than authoritative} cultural descriptions, especially for museum or ceremonial contexts.

Several limitations remain. The zero-shot setting is \textbf{province-level}, not token-level, so cultural term recovery depends on \textbf{retrieval-guided transfer} from seen-province captions. Retrieval may also propagate visually similar but culturally incorrect terminology. In addition, comparisons remain asymmetric between off-the-shelf and domain-trained models; future work will include fine-tuned BLIP and InstructBLIP baselines together with broader expert-based evaluation.

\section{Conclusion}

This paper proposes Custom ZeroCLIP, a retrieval-augmented vision-language framework for inductive zero-shot captioning of traditional Indonesian garments. Experimental and ablation results show that retrieval-guided domain adaptation is the main contributor to recovering province-specific garment terminology in unseen categories.

The findings suggest that retrieval-augmented multimodal systems can support scalable cultural heritage documentation in low-resource settings, including museum digitization, cultural indexing, and digital archiving, provided expert validation and cultural sensitivity are maintained.

\section*{Acknowledgements}

The authors thank the Intelligent Systems Laboratory, Faculty of Computer Science, Brawijaya University, for computational support, and the Indonesian traditional garment experts for their invaluable domain expertise.

\bibliographystyle{IEEEbib}
\bibliography{icme2026references}

\end{document}